\documentclass[runningheads]{llncs}
\usepackage{graphicx}

\usepackage{tikz}
\usepackage{comment}
\usepackage{amsmath,amssymb} 
\usepackage{color}

\usepackage[pagebackref,breaklinks,colorlinks]{hyperref}

\usepackage[accsupp]{axessibility}  

\usepackage[symbol]{footmisc}

\usepackage[nocompress]{cite}
\usepackage{mathrsfs}
\usepackage{multirow}
\usepackage{enumitem}
\usepackage{colortbl}
\usepackage{booktabs} 
\usepackage[misc]{ifsym}
\def\framework{\textit{Relighting4D}}
\definecolor{rred}{RGB}{245, 152, 153}
\definecolor{oorange}{RGB}{253, 205, 154}
\definecolor{yyellow}{RGB}{248,244,140}

\begin{document}
\pagestyle{headings}
\mainmatter
\def\ECCVSubNumber{34}  

\title{Relighting4D: \\ Neural Relightable Human from Videos} 

\titlerunning{Relighting4D: Neural Relightable Human from Videos}
%
\author{Zhaoxi Chen \and Ziwei Liu\thanks{Corresponding author.} }
\institute{S-Lab, Nanyang Technological University \\
\email{\small \{zhaoxi001, ziwei.liu\}@ntu.edu.sg}
}
\authorrunning{Z. Chen et al.}
\maketitle

\begin{abstract}
Human relighting is a highly desirable yet challenging task. Existing works either require expensive one-light-at-a-time (OLAT) captured data using light stage or cannot freely change the viewpoints of the rendered body. In this work, we propose a principled framework, \textbf{Relighting4D}, that enables free-viewpoints relighting from only human videos under unknown illuminations. Our key insight is that the space-time varying geometry and reflectance of the human body can be decomposed as a set of neural fields of normal, occlusion, diffuse, and specular maps. These neural fields are further integrated into reflectance-aware physically based rendering, where each vertex in the neural field absorbs and reflects the light from the environment. The whole framework can be learned from videos in a self-supervised manner, with physically informed priors designed for regularization. Extensive experiments on both real and synthetic datasets demonstrate that our framework is capable of relighting dynamic human actors with free-viewpoints. Codes are available at \url{https://github.com/FrozenBurning/Relighting4D}.

\keywords{Neural Rendering \and Dynamic Scenes \and Inverse Rendering }

\end{abstract}

\begin{figure}[t]
\centering
  \includegraphics[width=\textwidth]{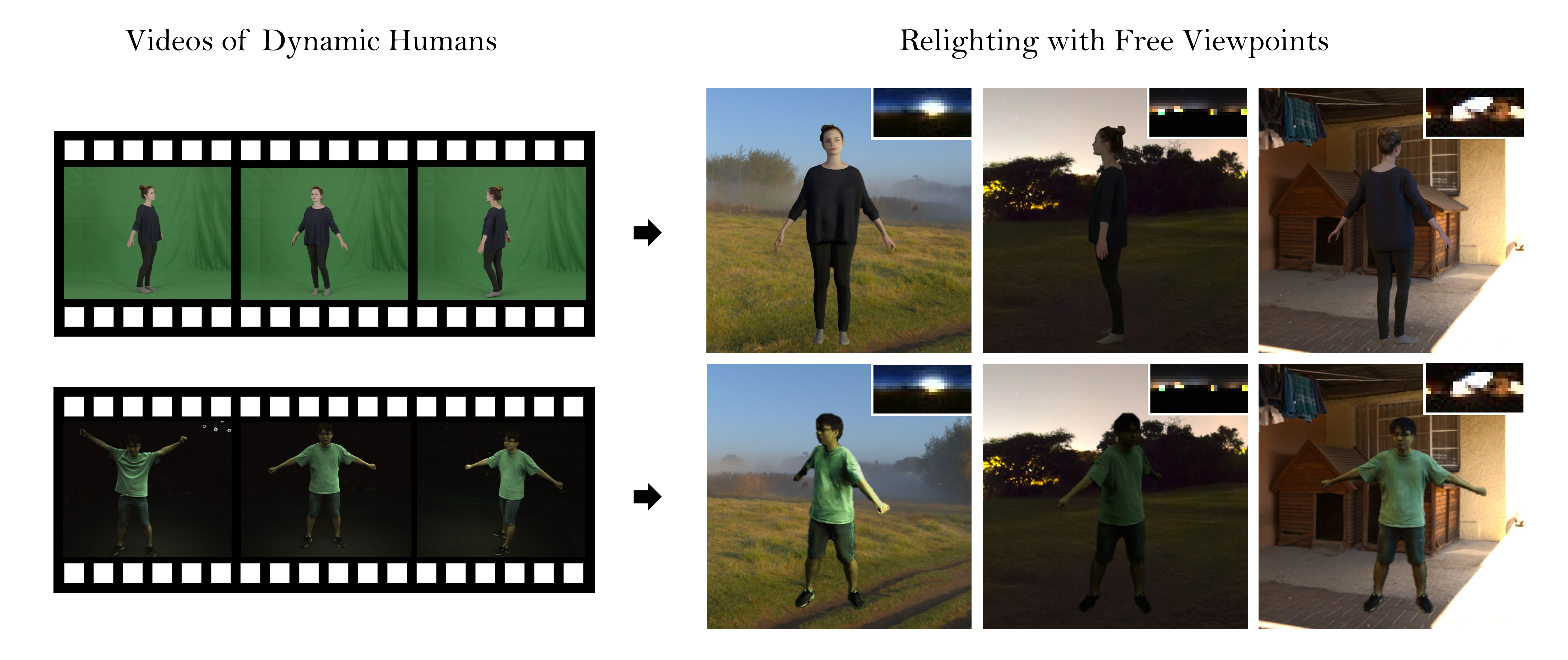}
  \caption{\textbf{Relighting of dynamic humans with free viewpoints}. \framework \;takes only videos as input, decomposing them into geometry and reflectance, which enables relighting of dynamic humans with free viewpoints by a physically based renderer.}
  \label{fig:fig1}
\end{figure}

\section{Introduction}
\label{sec:intro}

The emergence of metaverse has fueled the demands for photorealistic rendering of human characters, which benefits applications like digital 3D human and virtual reality. Among all factors, lighting is the most crucial one for rendering quality. Recently, remarkable success in relighting humans has been achieved~\cite{guo_relightables_2019,pandey_total_2021,zhang_neural_2021-1,zhang_neural_2021,meka_deep_2019,meka_deep_2020,bi_deep_2021,legendre_learning_2020,wang_single_2020,sun_nelf_2021}. However, the impressive quality of these methods heavily relies on the data captured by Light Stage~\cite{debevec_acquiring_2000}. The complicated hardware setup makes relighting systems expensive and only applicable in the constrained environment. On the other hand, a number of recent works propose to relight human images from a perspective of inverse rendering~\cite{tajima_relighting_2021,kanamori_relighting_2019,shu_neural_2017,zhou_deep_2019,liu_relighting_2021,lagunas_single-image_2021}. They succeed in relighting 2D images, yet fail to relight with novel views. A lack of underlying 3D representations impedes their flexibility of application.

In this paper, we focus on the problem of relighting dynamic humans from only videos, as illustrated in Figure \ref{fig:fig1}. The setting significantly reduces the cost of a flexible relighting system and broadens its scope of application. It has been proved~\cite{mildenhall_nerf_2020, sitzmann_scene_2020, pumarola_d-nerf_2020, zhang_editable_2021, ost_neural_2021, zhang_nerfactor_2021, srinivasan_nerv_2020, boss_nerd_2021, bi_neural_2020, martin-brualla_nerf_nodate, xian_space-time_2021, peng_neural_2021, park_nerfies_2021, li_neural_2021} that a scene can be represented as neural fields to enable novel view synthesis and relighting.
Among above methods, some~\cite{zhang_nerfactor_2021, srinivasan_nerv_2020, boss_nerd_2021, bi_neural_2020, martin-brualla_nerf_nodate, zhang_physg_2021} deal with relighting static objects but fail to model dynamic scenes. In sum, none of those methods successfully incorporate illuminations and scene dynamics simultaneously. 

Different from existing methods on novel view synthesis of the human body that are either non-relightable or require expensive OLAT captured images, we seek to estimate plausible geometry and reflectance from posed human videos.

To this end, we propose \textbf{\framework}, to relight dynamic humans with free viewpoints from videos given the 4D coordinates $(x,y,z,t)$ and the desired illumination. Specifically, our method first aggregates observations from posed human videos through space and time by a neural field conditioned on a deformable human model. Then, we decompose the neural field into geometry and reflectance counterparts, namely normal, occlusion, diffuse, and specular maps, which drive a physically based renderer to perform relighting.

We evaluate our approach on both monocular and multi-view videos. Overall, \framework \;outperforms other methods on perceptual quality and physical correctness. It relights dynamic humans in high fidelity, and generalizes to novel views. Furthermore, we demonstrate our capability of relighting under novel illuminations, especially the challenging OLAT setting, by creating a synthetic dataset called BlenderHuman for quantitative evaluations. 

We summarize our contributions as follows:
\textbf{1)} We present a principled framework, \framework, which is the first to relight dynamic humans with free viewpoints using only videos.
\textbf{2)} We propose to disentangle reflectance and geometry from input videos under unknown illuminations by leveraging multiple physically informed priors in a physically based rendering pipeline. 
\textbf{3)} Extensive experiments on both synthetic and real datasets demonstrate the feasibility and significant improvements of our approach over prior arts.

\section{Related Work}
\label{sec:review}

\noindent\textbf{Neural scene representation}~\cite{li_mine_nodate, tucker_single-view_2020, sitzmann_implicit_2020, sitzmann_scene_2020, mildenhall_nerf_2020, yu_pixelnerf_nodate, peng_neural_2021, pumarola_d-nerf_2020, liu_neural_2021-1, sun_neural_2021, suo_neuralhumanfvv_2021, kwon_neural_2021, raj_anr_nodate} has witnessed significant progress in representing a 3D scene with deep neural networks. NeRF~\cite{mildenhall_nerf_2020} proposes to model the scene as a 5D radiance field. To model dynamic humans, Neural Body~\cite{peng_neural_2021} proposes to attach a set of latent codes to a deformable human body model (i.e., SMPL~\cite{SMPL:2015}).
However, these methods implicitly incorporate all color information in the radiance field, which impedes their application towards relighting a dynamic human.

\noindent\textbf{Inverse rendering} aims to disentangle the appearance from observed images into geometry, material, and lighting condition. Previous works~\cite{meka_real-time_2021, legendre_deeplight_nodate, yu_inverserendernet_2018, wang_learning_nodate, sengupta_neural_2019, liu_relighting_2021, barron_shape_2012, liu_unsupervised_2020} seek to address it by conditioning on physically based priors or synthetic data. However, they fail in novel view synthesis due to the lack of underlying 3D representations. Recently, NeRF based methods~\cite{bi_neural_2020,zhang_nerfactor_2021, srinivasan_nerv_2020, boss_nerd_2021, zhang_physg_2021} propose to learn 3D reflectance fields or light transport fields from input images to enable free-viewpoint relighting. However, none of them is applicable to relight dynamic humans with space-time varying features.

\noindent\textbf{Relighting of human} face, avatar and body has wide-range applications~\cite{shu_neural_2017, sun_single_2019, wang_single_2020, zhou_deep_2019, pandey_total_2021}. As for full-body human relighting, convolutional methods~\cite{kanamori_relighting_2019, tajima_relighting_2021} fail to relight from novel viewpoints as there is no underlying 3D representation. Other methods~\cite{zhang_neural_2021-1, guo_relightables_2019} heavily relies on one-light-at-a-time~\cite{debevec_acquiring_2000} (OLAT) images, which is neither cheap to capture nor publicly available. \framework \;differentiates itself from aforementioned methods in that we achieve free-viewpoint relighting of dynamic full-body humans without the requirement on expensive capture setup.

\begin{figure}[t]
    \begin{center}
    \centerline{\includegraphics[width=\columnwidth]{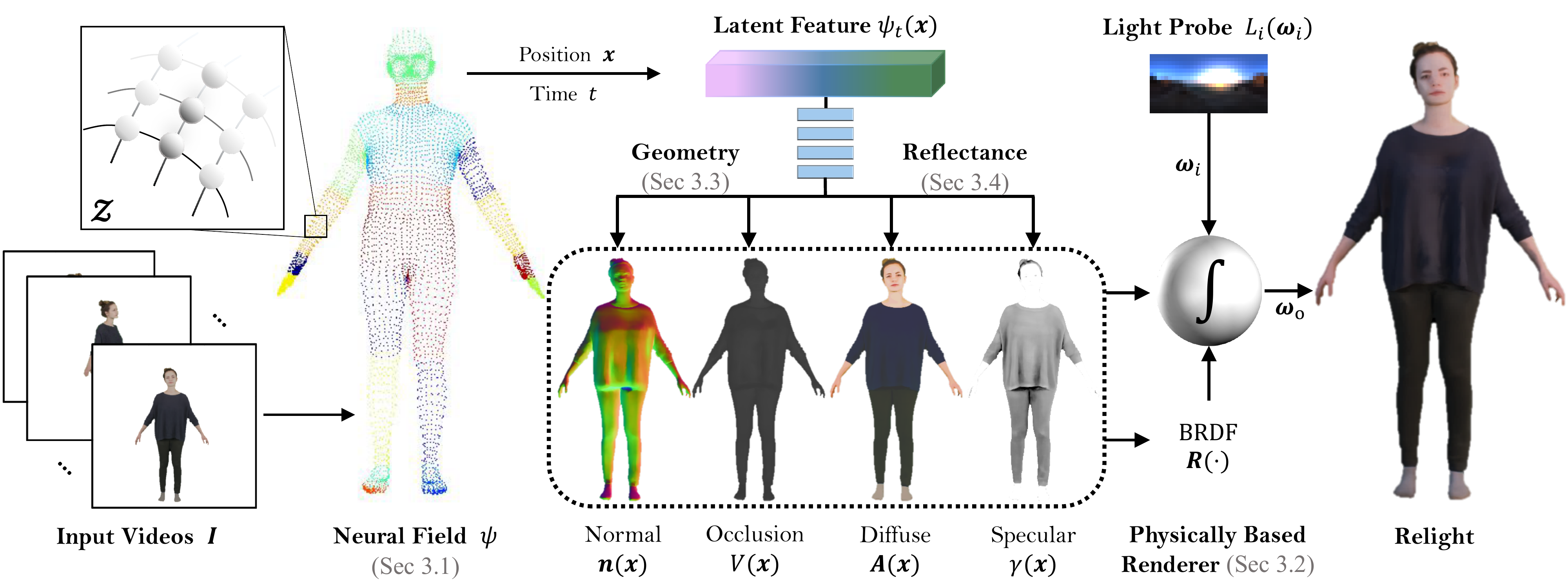}}
    \caption{\textbf{Overview of \framework}. Given the input video frame at time step $t$, \framework \;represents the human as a neural field $\psi$ on latent vectors $\mathcal{Z}$ anchored to a deformable human model. The value of the neural field $\psi_t(\boldsymbol{x})$ at any 3D point $\boldsymbol{x}$ and time $t$ is taken as latent feature and fed into multilayer perceptrons to obtain geometry and reflectance, which are normal, occlusion, diffuse, and specular maps respectively. Finally, a physically based renderer is raised to render the human subject to the input light probe under novel illumination.
    }\label{fig:overview}
    \end{center}
\end{figure}

\section{Our Approach}
\label{sec:method}
Given a human video, \framework \;can synthesize videos with free viewpoints under novel illuminations. We denote the input video as $I = \{I_1, I_2, ..., I_t\}$, where $t$ is the time step. In general, our model learns a physically based renderer from $I$. During inference, it takes a 3D position $\boldsymbol{x} \in \mathbb{R}^3$, a time step $t$, a camera view $\boldsymbol{\omega}_o \in \mathbb{R}^3$, a desired light probes $\boldsymbol{L}_i \in \mathbb{R}^{16\times32\times3}$ as inputs, and outputs the corresponding outgoing radiance $\boldsymbol{L}_o \in \mathbb{R}^3$.

\noindent\textbf{Framework overview.} We first give an overview of \framework~(Figure \ref{fig:overview}). It first derives latent features from the video, which is achieved by estimating a neural field.
Based on the latent features, \framework \;decomposes the human performer into geometry and reflectance information which drive our physically based renderer. The space-time varying geometry and reflectance of the full human body are parameterized by four multilayer perceptrons. Note that, \framework \;enables relighting of dynamic humans with free viewpoints using \textbf{only} videos, without training on any captured data (e.g., OLAT or flash images).

\subsection{Neural Field as Human Representation}
\label{sec:latent}
Extracting 4D representations of dynamic human performers is a non-trivial task. Compared to the static scenes where NeRF~\cite{mildenhall_nerf_2020} fits well, dynamic scenes in videos have factors like motion, occlusion, non-rigid deformation, and illumination that vary through space and time which hampers an accurate estimation. 

Inspired by the local implicit representations~\cite{peng_neural_2021, genova_local_2020},
we introduce a 4D neural field $\psi$ conditioned on a parametric human model (SMPL~\cite{SMPL:2015} or SMPL-X~\cite{SMPL-X:2019}) to represent a dynamic human performer, which maps the position $\boldsymbol{x}$ and time step $t$ to the latent feature $\psi_t(\boldsymbol{x})$. 
Specifically, at frame $I_t$, we obtain the parameters of human model (i.e. locations of vertices) using this tool~\cite{total_capture}. Then, a set of latent vectors $\mathcal{Z} \in \mathbb{R}^{N\times16}$ is assigned to the vertices of human model, where $N=6890$ for SMPL~\cite{SMPL:2015} and $N=10475$ for SMPL-X~\cite{SMPL-X:2019}. Then we query the neural field by the 4D coordinates $(\boldsymbol{x}, t)$, extracting the latent feature $\psi_t(\boldsymbol{x}) \in \mathbb{R}^{256}$ from $\mathcal{Z}$ via trilinear interpolation of its nearby vertices.

NeuralBody~\cite{peng_neural_2021} employs a similar strategy on human representations. But it's not relightable in the way that it fails to disentangle geometry and reflectance from the latent codes. In contrast, \framework \;learns a distinct neural field that can be decomposed into geometry (Section \ref{sec:geo}) and reflectance (Section \ref{sec:refl}), which serves the physically based renderer (Section \ref{sec:pbr}) for relighting.

\subsection{Physically Based Rendering}
\label{sec:pbr}
\begin{figure}[t]
\begin{minipage}{0.55\columnwidth}
    \begin{center}
    \centerline{\includegraphics[width=1\columnwidth]{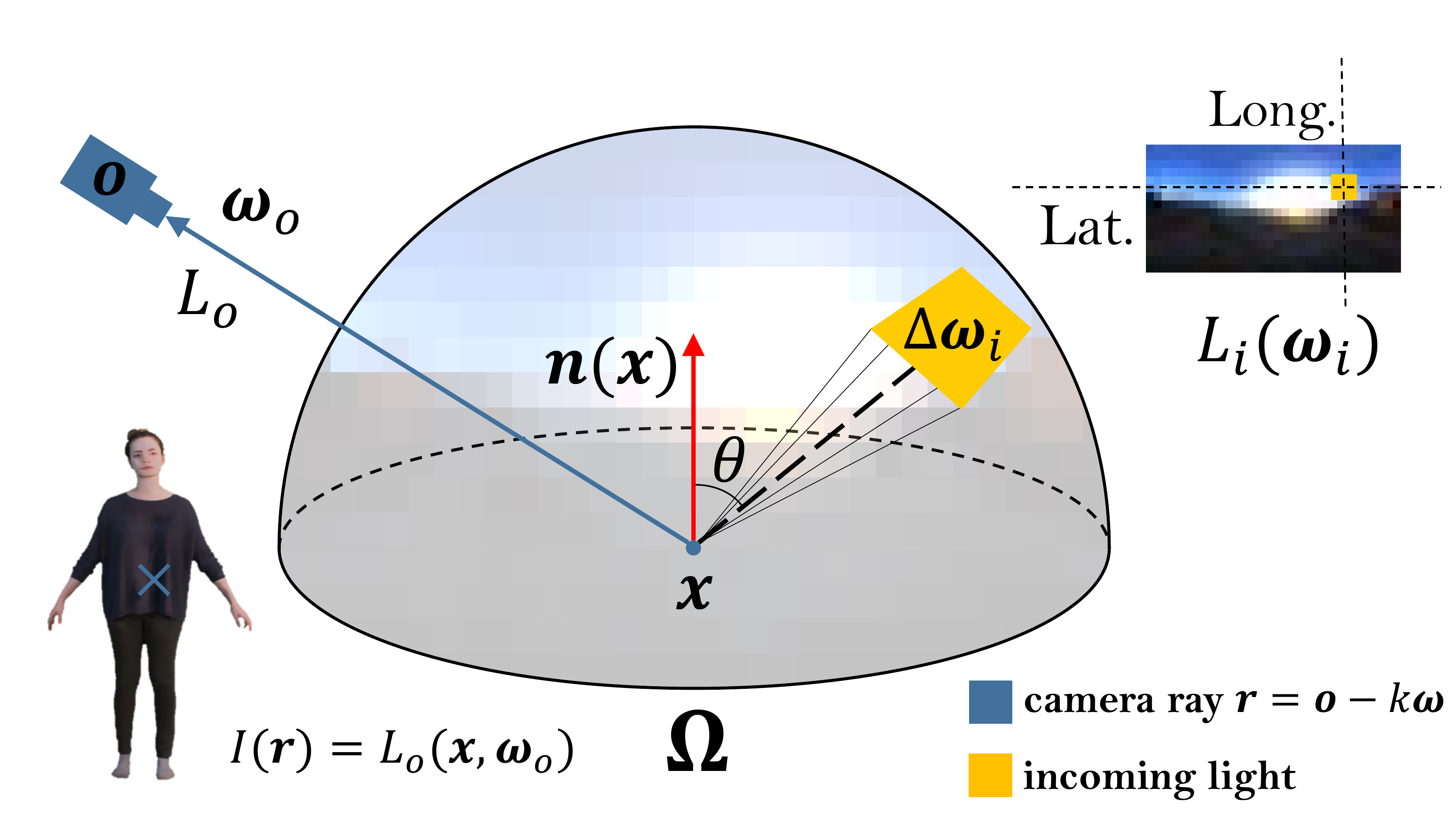}}
    \vskip -0.05in
    \caption{\textbf{Illustration of our physically based rendering pipeline}. The environment light is represented as spherical coordinates in latitude-longtitude (Lat.-Long.) format. Given the surface location $\boldsymbol{x}$, the incoming light from $\boldsymbol{\omega}_i$ with the area of $\Delta \boldsymbol{\omega}_i$ is scattered by the microfacet that is parameterized by BRDF $R(\cdot)$,
    normal $\boldsymbol{n}(\boldsymbol{x})$, and $\cos{\theta} = \boldsymbol{\omega}_i \cdot \boldsymbol{n}(\boldsymbol{x})$. Then the outgoing radiance $L_o(\boldsymbol{x}, \boldsymbol{\omega}_o)$ along the ray $\boldsymbol{r} = \boldsymbol{o} - k\boldsymbol{\omega}_o$ is calculated according to Eqn. \ref{eq:render-discrete}, which equals to the corresponding pixel value.
    }\label{fig:pbr}
    \end{center}
\end{minipage}
\hspace{0.1in}
\begin{minipage}{0.44\columnwidth}
    \vskip 0.2in
    \begin{center}
    \centerline{\includegraphics[width=1\columnwidth]{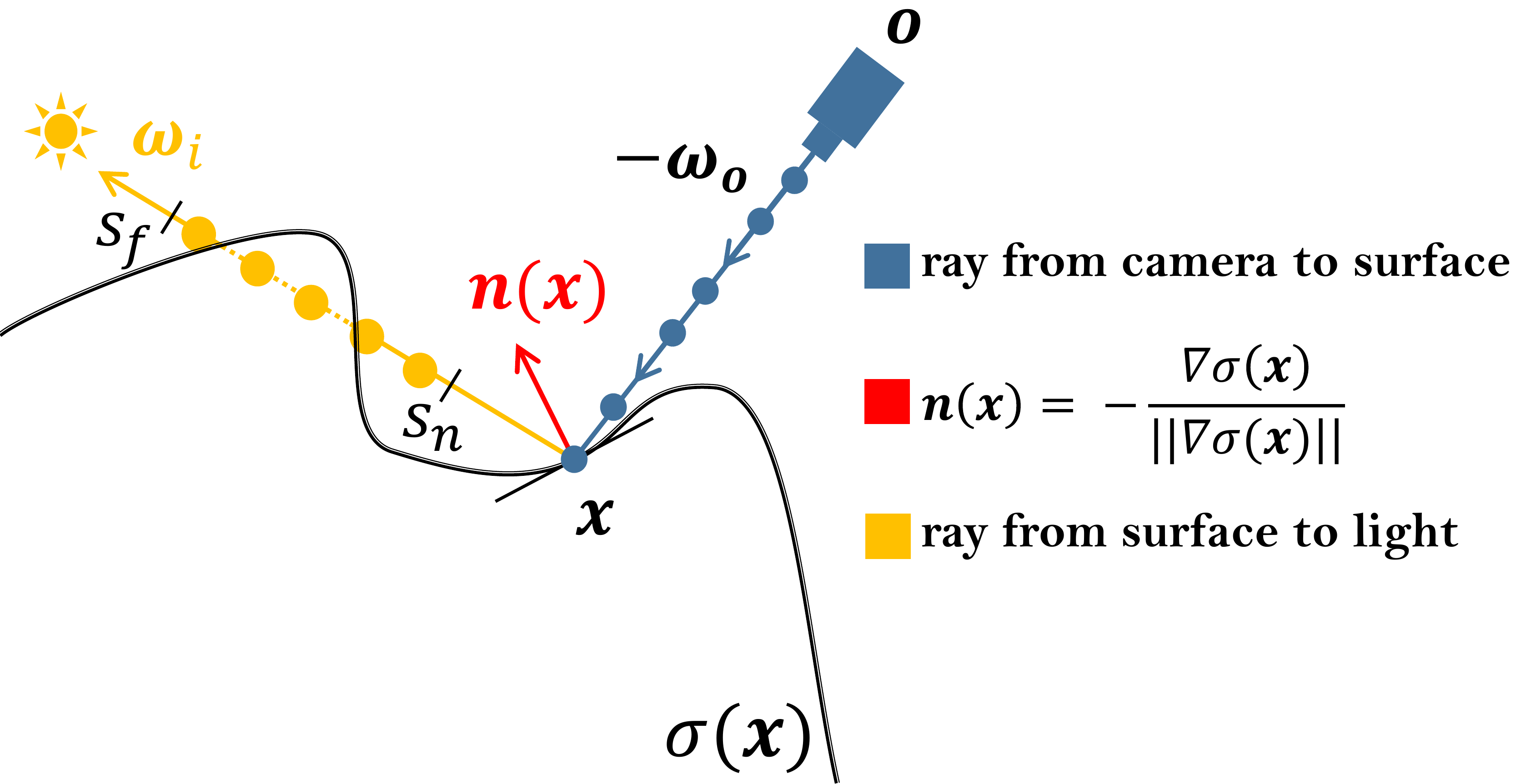}}
    \vskip 0.15in
    \caption{\textbf{The process of baking geometry}. Note that we perform two different types of ray marching during training. The one is marching along the camera ray $\boldsymbol{r} = \boldsymbol{o} - k\boldsymbol{\omega}_o$ to the expected depth of termination $k$ to get the geometry surface $\boldsymbol{x}$, while the other is marching from the surface $\boldsymbol{x}$ to the light coming from direction $\boldsymbol{\omega}_i$ to calculate the accumulated transmittance (occlusion map). 
    }\label{fig:geo}
    \end{center}
\end{minipage}
\end{figure}

While differentiable volume rendering has been used in recent works~\cite{mildenhall_nerf_2020, yu_pixelnerf_nodate, peng_neural_2021}, these methods focus on novel view synthesis with radiance fields. In general, to enable relighting with neural representations, instead of modeling the human body as a field of vertices that \textit{emit} light, we represent the human as a field of vertices that \textit{reflect} the light from the environment. Specifically, we leverage a physically based renderer, which models a reflectance-aware rendering process. Mathematically, our rendering pipeline is driven by the following equation:
\begin{equation}
\label{eq:render-continuous}
\small
    L_o(\boldsymbol{x}, \boldsymbol{\omega}_o) = \int_{\Omega}R(\boldsymbol{x}, \boldsymbol{\omega}_i, \boldsymbol{\omega}_o, \boldsymbol{n}(\boldsymbol{x}))L_i(\boldsymbol{x},\boldsymbol{\omega}_i)(\boldsymbol{\omega}_i \cdot \boldsymbol{n}(\boldsymbol{x})) d \boldsymbol{\omega}_i,
\end{equation}
where $L_o(\boldsymbol{x},\boldsymbol{\omega}_o) \in \mathbb{R}^3$ is the outgoing radiance at point $\boldsymbol{x}$ viewed from $\boldsymbol{\omega}_o$. $L_i(\boldsymbol{x},\boldsymbol{\omega}_i) \in \mathbb{R}^3$ is the incident radiance arriving at $\boldsymbol{x}$ from direction $\boldsymbol{\omega}_i$. $\Omega$ is an unit sphere that models all possible light directions, and $\boldsymbol{n}(\boldsymbol{x}) \in \mathbb{R}^3$ is the normal. $R(\boldsymbol{x}, \boldsymbol{\omega}_i, \boldsymbol{\omega}_o, \boldsymbol{n}(\boldsymbol{x}))$\footnote[2]{For simplicity, we also use $R(\cdot)$ to denote BRDF when necessary.} is the Bidirectional Reflectance Distribution Function (BRDF) which defines how the incident light is reflected at the surface, and $d\boldsymbol{\omega}_i$ is the solid angle of incident light at $\boldsymbol{\omega}_i$. We use a discrete set of light samples to approximate Eqn. \ref{eq:render-continuous} in the following way:
\begin{equation}
\label{eq:render-discrete}
\small
    L_o(\boldsymbol{x}, \boldsymbol{\omega}_o) \approx \sum_{\boldsymbol{\omega}_i}R(\boldsymbol{x}, \boldsymbol{\omega}_i, \boldsymbol{\omega}_o, \boldsymbol{n}(\boldsymbol{x}))L_i(\boldsymbol{x},\boldsymbol{\omega}_i)(\boldsymbol{\omega}_i \cdot \boldsymbol{n}(\boldsymbol{x})) \Delta \boldsymbol{\omega}_i,
\end{equation}
where $\Delta \boldsymbol{\omega}_i$ is sampled from a light probe that depicts the distribution of light sources in space. We represent the environment light $L_i(\boldsymbol{\omega}_i)$ as a light probe image in latitude-longitude format with a resolution of $16\times 32\times3$, which facilitates relighting applications by replacing the estimated light probe with an external one. 
Figure \ref{fig:pbr} illustrates our physically based renderer at surface $\boldsymbol{x}$.

Note that previous work~\cite{mildenhall_nerf_2020, peng_neural_2021} implicitly encodes $R(\cdot)$ in the radiance fields without modeling the reflectance. To enable flexible relighting applications, we leverage the microfacet model~\cite{Walter:ea:2007} to approximate a differentiable reflectance function parameterized by the surface normal $\boldsymbol{n}(\boldsymbol{x})$, the diffuse map $\boldsymbol{A}(\boldsymbol{x})$ and the specular roughness $\gamma(\boldsymbol{x})$. Due to the limited space, we introduce the implementation of $R(\cdot)$ in the supplementary.

To encode harsh shadow and occlusion, we mask the incident light $L_i(\boldsymbol{x},\boldsymbol{\omega}_i)$ by the occlusion map $V(\boldsymbol{x},\boldsymbol{\omega}_i)$ at $\boldsymbol{x}$:
\begin{equation}
\label{eq:occlusion}
\small
    L_i(\boldsymbol{x},\boldsymbol{\omega}_i) = V(\boldsymbol{x},\boldsymbol{\omega}_i)L_i(\boldsymbol{\omega}_i).
\end{equation}

\noindent\textbf{Physical characteristics disentanglement.} 
Driven by Eqn. \ref{eq:render-discrete}, the renderer requires physical characteristics, i.e., geometry, reflectance, and light, of a given human performer, which are disentangled and estimated by \framework \;from input videos. The details are introduced in the following two sections.

\subsection{Volumetric Geometry}
\label{sec:geo}
In terms of geometry, our renderer requires a normal map $\boldsymbol{n}(\boldsymbol{x}) \in \mathbb{R}^3$ and an occlusion map $V(\boldsymbol{x}, \boldsymbol{\omega}_i) \in \mathbb{R}$ as inputs. 
Moreover, we render on the surface to keep the computing process tractable, which requires an estimation of surface position. 
It can be easily obtained by querying a density field. 

We first reconstruct the geometry of the given scene using an auxiliary density field $f_\sigma: (\boldsymbol{x}, \psi_t(\boldsymbol{x})) \xrightarrow{} \sigma(\boldsymbol{x})$. It's derived from the latent feature $\psi_t(\boldsymbol{x})$ using an MLP. As shown in Figure \ref{fig:geo}, \framework \;leverages the auxiliary density field by baking it into surface maps, normal maps and occlusion maps.

\textbf{Surface map} is the 3D coordinates of points at the expected termination of depth given the camera view $\boldsymbol{\omega}_o$. We march the camera ray $\boldsymbol{r}$ from its origin $\boldsymbol{o}$ along the direction $-\boldsymbol{\omega}_o$ to the expected termination of depth $k$ to get the surface $\boldsymbol{x} = \boldsymbol{o} - k\boldsymbol{\omega}_o$.

\textbf{Normal map} is computed on the surface as the normalized negative gradient of the density field: $\Tilde{\boldsymbol{n}}(\boldsymbol{x}) = -\nabla \sigma(\boldsymbol{x})/||\nabla \sigma(\boldsymbol{x})||$.

\textbf{Occlusion map} denotes the transmittance of surface points from a specific direction. We compute the occlusion map by marching the ray $\boldsymbol{r}(s,\boldsymbol{x},\boldsymbol{\omega}_i) = \boldsymbol{x} + s\boldsymbol{\omega}_i$ from the surface of the human body to the corresponding light at $\boldsymbol{\omega}_i$: $\Tilde{V}(\boldsymbol{x}, \boldsymbol{\omega}_i) = 1 - exp(-\int^{s_f}_{s_n} \sigma(\boldsymbol{r}(s, \boldsymbol{x}, \boldsymbol{\omega}_i))ds)$, where $s_n$ and $s_f$ is the near and far bounds along the direction of the light. We set $s_n=0, s_f=0.5$ for all scenes. In other words, occlusion map considers the visibility at the given surface $\boldsymbol{x}$ by querying the density fields from $s_n$ to $s_f$ along the incident light direction $\boldsymbol{\omega}_i$.

Unfortunately, directly using the baked geometry causes numerous queries of $f_\sigma$(e.g., for occlusion map, we should trace $16\times32 = 512$ rays from all possible lighting directions for one 3D point), which is not tractable  during training and rendering. Thus, we use an MLP $f_n: (\boldsymbol{x}, \psi_t(\boldsymbol{x})) \xrightarrow{} \boldsymbol{n}(\boldsymbol{x})$ to reparameterize the surface and latent features to the normal map, and another MLP $f_V: (\boldsymbol{x}, \omega_i, \psi_t(\boldsymbol{x})) \xrightarrow{} V(\boldsymbol{x})$ to map the surface, light direction and features to the occlusion map $V$. 
The weights of $f_V, f_n$ are trained with the geometry reconstruction loss, intending to recover the baked geometry:
\begin{equation}
\small
    \mathcal{L}_{geo} = ||V(\boldsymbol{x}) - \Tilde{V}(\boldsymbol{x})||^2_2 + ||\boldsymbol{n}(\boldsymbol{x}) - \Tilde{\boldsymbol{n}}(\boldsymbol{x})||^2_2.
\end{equation}

\noindent\textbf{Smoothness regularization.}
We regularize $f_V, f_n$ by L1 penalty to keep the smoothness of their outputs:
\begin{equation}
\label{eq:geo-smoothness}
\small
    \tau_V = |V(\boldsymbol{x}) - V(\boldsymbol{x}+\boldsymbol{\epsilon})|_1 \quad \tau_n = |\boldsymbol{n}(\boldsymbol{x}) - \boldsymbol{n}(\boldsymbol{x}+\boldsymbol{\epsilon})|_1,
\end{equation}
where we measure the local smoothness by adding 3D perturbation $\boldsymbol{\epsilon}$ to $\boldsymbol{x}$ which is sampled from a Gaussian distribution with zero mean and standard deviation $0.01$.
Several works~\cite{oechsle_unisurf_2021,zhang_nerfactor_2021} have validated the use of similar smoothness losses for the aim of shape reconstruction. 

\noindent\textbf{Temporal coherence regularization.} It is crucial for a 4D representation to incorporate temporal coherence. Otherwise, the rendered sequence will contain jitter appearance. Moreover, an accurate geometry is also important for artifact-free physically based rendering. Therefore, we add the following regularization term to encourage a temporally smooth geometry:
\begin{equation}
\label{eq:corr}
    \mathcal{L}_{temp} = \frac{1}{N}\sum^N_{i=1}|\sigma_t(\hat{\boldsymbol{x}_i}) - \sigma_{t+1}(\hat{\boldsymbol{x}}_i)|_1,
\end{equation}
where $\hat{\boldsymbol{x}}_i$ is the 3D position of $i$-th vertex of SMPL model. Eqn. \ref{eq:corr} explicitly constrains the temporal coherence of the geometry, and also implicitly regularize the latent feature $\psi_t(\boldsymbol{x})$ which benefits the following reflectance estimation.

\subsection{Reflectance}
\label{sec:refl}
In terms of reflectance, our physically based renderer requires the BRDF $R(\cdot)$ and the light probe $L_i(\boldsymbol{\omega}_i)$ as inputs. As presented in Section \ref{sec:pbr}, our BRDF estimation consists of a Lambertian RGB diffuse component $\boldsymbol{A}(\boldsymbol{x}) \in \mathbb{R}^3$ and a specular component $\gamma(\boldsymbol{x}) \in \mathbb{R}$. We parameterize the diffuse map at $\boldsymbol{x}$ with latent features $\psi_t(\boldsymbol{x})$ as an MLP $f_A: (\boldsymbol{x}, \psi_t(\boldsymbol{x})) \xrightarrow{} \boldsymbol{A}(\boldsymbol{x})$, and parameterize the specular map as another MLP $f_\gamma: (\boldsymbol{x}, \psi_t(\boldsymbol{x})) \xrightarrow{} \gamma(\boldsymbol{x})$. 

\noindent\textbf{Local smoothness prior.} The problem that decomposes BRDF from video frames under unknown illumination is highly ill-posed. As the color information is entangled, and there is no off-the-shelf supervision on the reflectance. Inspired by work~\cite{barron_shape_2012, boss_nerd_2021, sengupta_neural_2019, zhang_nerfactor_2021, zhang_physg_2021} on intrinsic decomposition which leverages piece-wise smoothness prior on albedo, we regularize the optimization of $f_A$ by L1 penalty:
\begin{equation}
\label{eq:refl-smoothness}
\small
    \tau_A = |\boldsymbol{A}(\boldsymbol{x}) - \boldsymbol{A}(\boldsymbol{x}+\boldsymbol{\epsilon})|_1,
\end{equation}
where $\boldsymbol{\epsilon}$ is the same type of perturbation as Eqn. \ref{eq:geo-smoothness}.

\noindent\textbf{Global sparsity prior.} However, given this under-constrained problem, the local smoothness regularization in Eqn. \ref{eq:refl-smoothness} is not sufficient for a plausible estimation of the diffuse map, as shown in Figure \ref{fig:albedo-entropy}. Thus, we further leverage global minimum-entropy sparsity prior on diffuse map which has been previously explored\cite{bas-relief,DBLP:journals/ijcv/FinlaysonDL09, prior, barron_shape_2012} on shadow removal. From a perspective of physically based rendering, the diffuse map represents the base color, indicating that the palette should be sparse enough. Intuitively, the diffuse map of clothes should contains a small number of colors. Thus, we minimize the Shannon entropy of diffuse map, denoted as $H_{A}$, to impose this prior on our model. Since the diffuse map $\boldsymbol{A}(\boldsymbol{x})$ is a continuous variable whose probability density function (PDF) is unknown, a naive way to estimate its entropy is using histogram to get PDF. But, it's not differentiable. Instead, it's always possible to use a soft and differentiable generalization of Shannon entropy (i.e. quadratic entropy\cite{mutual-info}). However, it's quadratically expensive to the number of sampled camera rays.

This motivates our novel approximation of minimizing $H_{A}$ in a both differentiable and efficient way. The key insight is that the PDF of $\boldsymbol{A}(\boldsymbol{x})$, $p(\boldsymbol{A}(\boldsymbol{x}))$, can be estimated by a Gaussian KDE (Kernel Density Estimator). Given a diffuse map $\boldsymbol{A}(\boldsymbol{x})$, we leverage a KDE as its PDF approximation:
\begin{equation}
\label{eq:kde}
    \Tilde{p}(\boldsymbol{A}(\boldsymbol{x})) =  \frac{1}{n}\sum^{n}_{i=1} K_G(\boldsymbol{A}(\boldsymbol{x})-\boldsymbol{A}_i(\boldsymbol{x})),
\end{equation}
where $K_G$ is the standard normal density function, $n$ is the number of sampled rays during training, and $\boldsymbol{A}_i(\boldsymbol{x})$ is the value of diffuse map at the $i$-th camera ray. Then the entropy of $\boldsymbol{A}(\boldsymbol{x})$ is computed as an expectation:

\begin{equation}
    H_{A} = \mathbb{E}[-\log(\Tilde{p}(\boldsymbol{A}(\boldsymbol{x})))].
\end{equation}

In addition, as the input video is captured under unknown illuminations, we randomly initialize the light probe $L_i(\boldsymbol{\omega}_i)$ as a trainable parameter, optimizing it during the training phase to estimate a plausible ambient light of the scene. It can be replaced by a new HDR map for relighting after training.

\subsection{Progressive End-to-End Learning}
\label{sec:training}
In the training phase, we randomly sample 1024 camera rays for each input frame. Besides, we employ a progressive training strategy which allows the resolution of video to gradually increase. In specific, before ray sampling, the input video is scaled to the resolution of $\alpha H\times \alpha W$ where $\alpha \in (0, 1]$ is a monotonically increasing function of the number of iterations.

Furthermore, we embed the surface position $\boldsymbol{x}$ and the light direction $\boldsymbol{\omega}_i$ using the positional encoding~\cite{mildenhall_nerf_2020, tancik_fourier_2020} before concatenating them with latent features $\psi_t(\boldsymbol{x})$. The maximum frequency of set to $2^{10}$ and $2^4$, respectively. 
We use four fully-connected ReLU layers with 256 channels for each MLP.

Our full loss function is a summation:
\begin{equation}
\label{eq:loss}
\small
    \mathcal{L} = \lambda_{rgb}\mathcal{L}_{rgb} + \lambda_{geo}\mathcal{L}_{geo} + \lambda_{temp}\mathcal{L}_{temp} + \lambda_V\tau_V + \lambda_n\tau_n + \lambda_A\tau_A + \lambda_H H_A,
\end{equation}
where $\mathcal{L}_{rgb}$ is the reconstruction loss against the ground-truth pixel color value. We train each model for 260k iterations with a Tesla V100 GPU.
Details of training hyperparameters are deferred to the supplementary.

\begin{figure}[htp]
    \begin{center}
    \centerline{\includegraphics[width=\columnwidth]{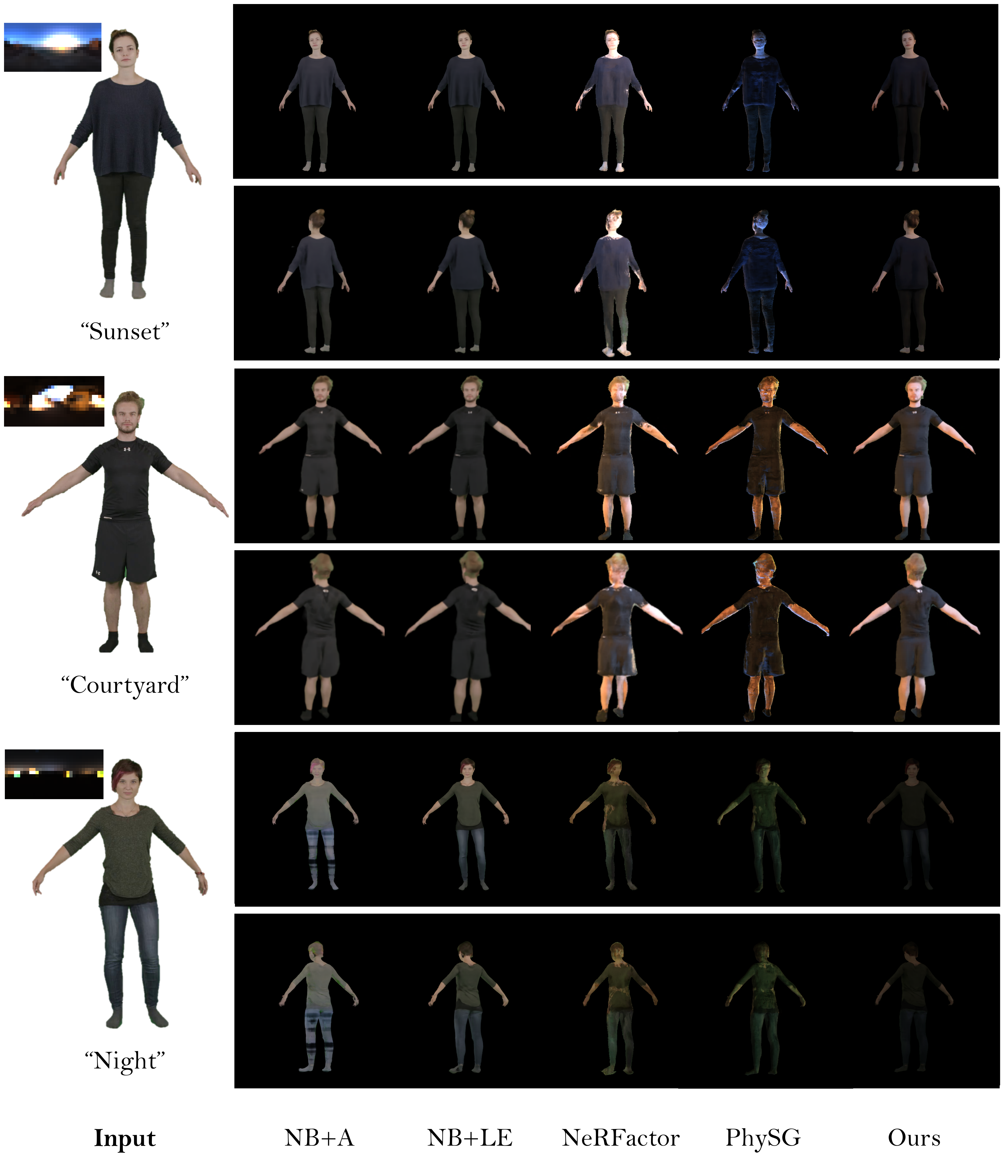}}
    \caption{\textbf{Free-viewpoint relighting on the People-Snapshot dataset}. Two variants of NeuralBody~\cite{peng_neural_2021} (NB+A and NB+LE) fail to incorporate the lighting in a physical way, thus are unable to reasonably relight the human actor. NB+A learns the wrong mapping between the target light and the appearance. And NB+LE reconstruct the input video well yet fails to generalize to novel lightings. NeRFactor~\cite{zhang_nerfactor_2021} and PhySG~\cite{zhang_physg_2021} seem to model physically correct illuminations but gives blurry results due to the incapability of modeling dynamics. \framework \;significantly outperforms comparison methods. \textbf{We show more results in both ambient lighting and OLAT setting in the supplementary videos.}
    }\label{fig:vis-snapshot}
    \end{center}
\end{figure}

\section{Experiments}
\label{sec:exp}
\noindent\textbf{Rendering settings.} We render humans in both the ambient lighting and the OLAT setting. For ambient lighting, we use publicly available\footnote[2]{\url{https://polyhaven.com/}} HDRi maps as light probes. Furthermore, for the OLAT setting, we simulate point lights by generating one-hot light probes given the incoming light directions.

\noindent\textbf{Real datasets.} We validate our method on the People-Snapshot~\cite{alldieck_video_2018} dataset and ZJU-Mocap~\cite{peng_neural_2021} dataset qualitatively. People-Snapshot~\cite{alldieck_video_2018} captures monocular videos with dynamic performers that keep rotating. And ZJU-Mocap~\cite{peng_neural_2021} captures dynamic humans with complex motions using a multi-camera system.

\noindent\textbf{Synthetic dataset.} To further demonstrate the effectiveness of \framework, we create a dataset, \textbf{BlenderHuman}, using the Blender engine~\cite{blender} for quantitative evaluation. Details will be deferred to the supplementary.

\noindent\textbf{Comparison methods.} 
We compare \framework \;with several competitive methods. \textbf{NeRFactor}~\cite{zhang_nerfactor_2021} requires a pretrained NeRF as a geometry proxy and learns a data-driven BRDF to perform relighting, but it fails to represent dynamic scenes. 
\textbf{PhySG}~\cite{zhang_physg_2021} adopts a spherical Gaussians reflectance model which cannot handle high-frequency lights, and its geometry representation cannot model dynamic scenes.
Moreover, to demonstrate the importance of physically based rendering, we implement two variants on top of NeuralBody (NB)~\cite{peng_neural_2021} which succeeds in novel view synthesis of dynamic humans but fails to incorporate lighting and reflectance. \textbf{NB+Ambient Light} (NB+A) uses a flattened light probe as the latent code which contributes to the prediction of its color model, while \textbf{NB+Learnable Embedding} (NB+LE) maps the light probe into a latent code using an MLP with two layers.

\noindent\textbf{Evaluation metrics.} For quantitative analysis, we use Peak Signal-to-Noise Ratio (PSNR), Structural Similarity Index Measure~\cite{ssim}(SSIM), and Learned Perceptual Image Patch Similarity~\cite{zhang2018perceptual}(LPIPS) as metrics. In addition, we use the error of degree($^\circ$) to measure the normal map estimations.

\subsection{Results on Real Datasets}

\noindent\textbf{Performance on relighting with novel views. } Figure \ref{fig:vis-snapshot} shows qualitative results on People-Snapshot dataset. All methods train a separate model for each human performer and re-render the human given the input light probes. Two variants of NeuralBody~\cite{peng_neural_2021}, NB+A and NB+LE, are good at reconstructing appearance but fail to incorporate novel illuminations in a perceptually salient way. They fail to learn the underlying physics of rendering. For example, NB+A maps the input light probe to artifacts of texture while NB+LE even seems to discard the features from lightings. NeRFactor~\cite{zhang_nerfactor_2021} and PhySG~\cite{zhang_physg_2021} give blurry results, which show that they cannot aggregate space-time varying geometry and reflectance of dynamic humans, leading to degraded rendering results. In contrast, our method generates photorealistic relit novel views.

We also present our qualitative results in the challenging OLAT setting. Since the point light comes from only one direction, these OLAT illuminations induce hard cast shadows, effectively revealing rendering artifacts due to inaccurate geometry and materials. \framework \;synthesizes shadows cast by limbs and clothes in a physically correct way. Please refer to the supplementary for details.

\begin{figure}[t]
    \begin{center}
    \begin{minipage}{0.48\columnwidth}
    \vskip 0.2in
    \centerline{\includegraphics[width=1.04\columnwidth]{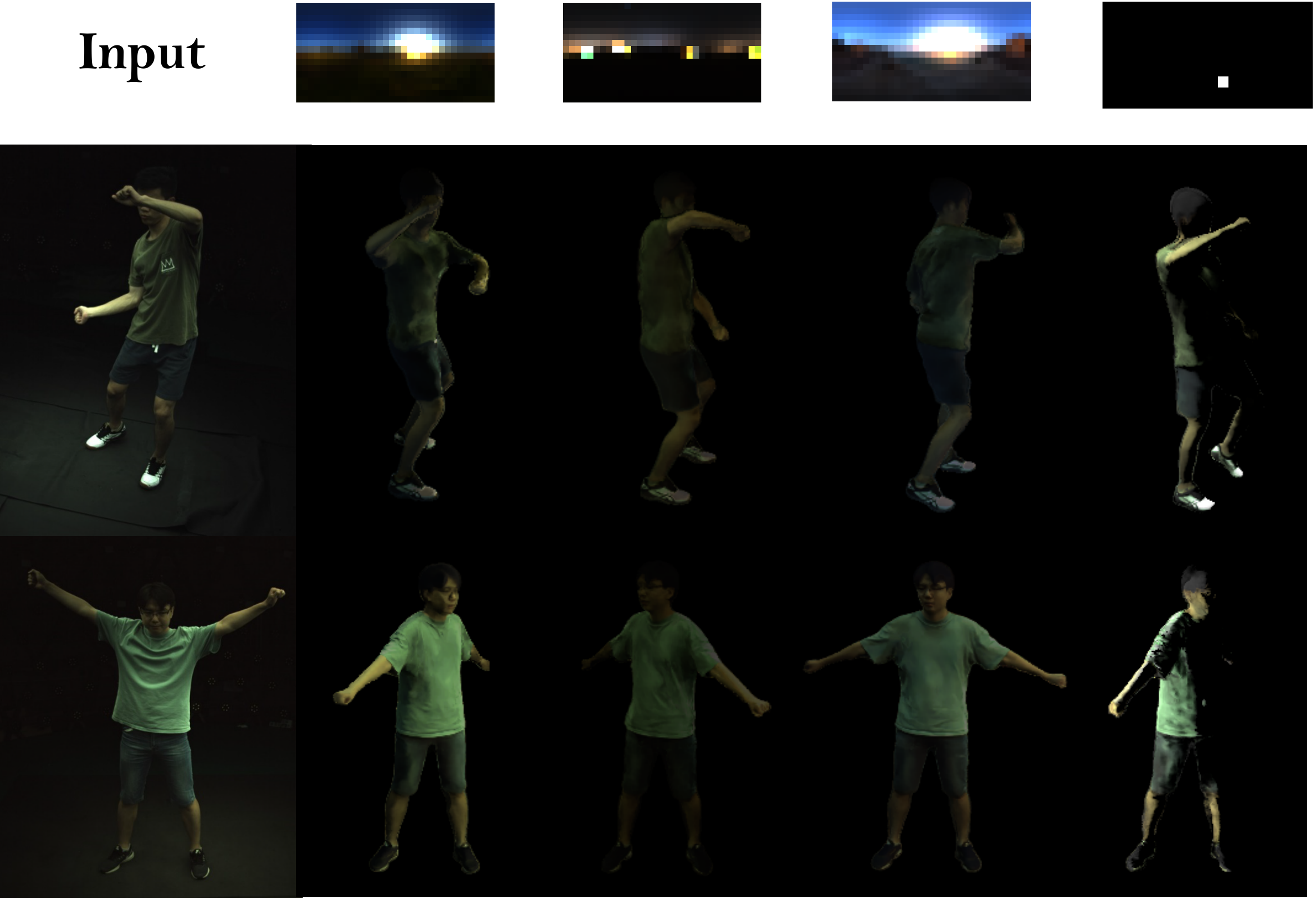}}
    \caption{\textbf{Relighting of dynamic humans with complex motions on ZJU-Mocap dataset}. \framework \;renders high-fidelity human actors with time-varying poses under novel illuminations. \textbf{Please check the supplementary videos for more results.}
    }\label{fig:vis-zju}
    \end{minipage}
    \hspace{0.1in}
    \begin{minipage}{0.48\columnwidth}
    \centerline{\includegraphics[width=1.0\columnwidth]{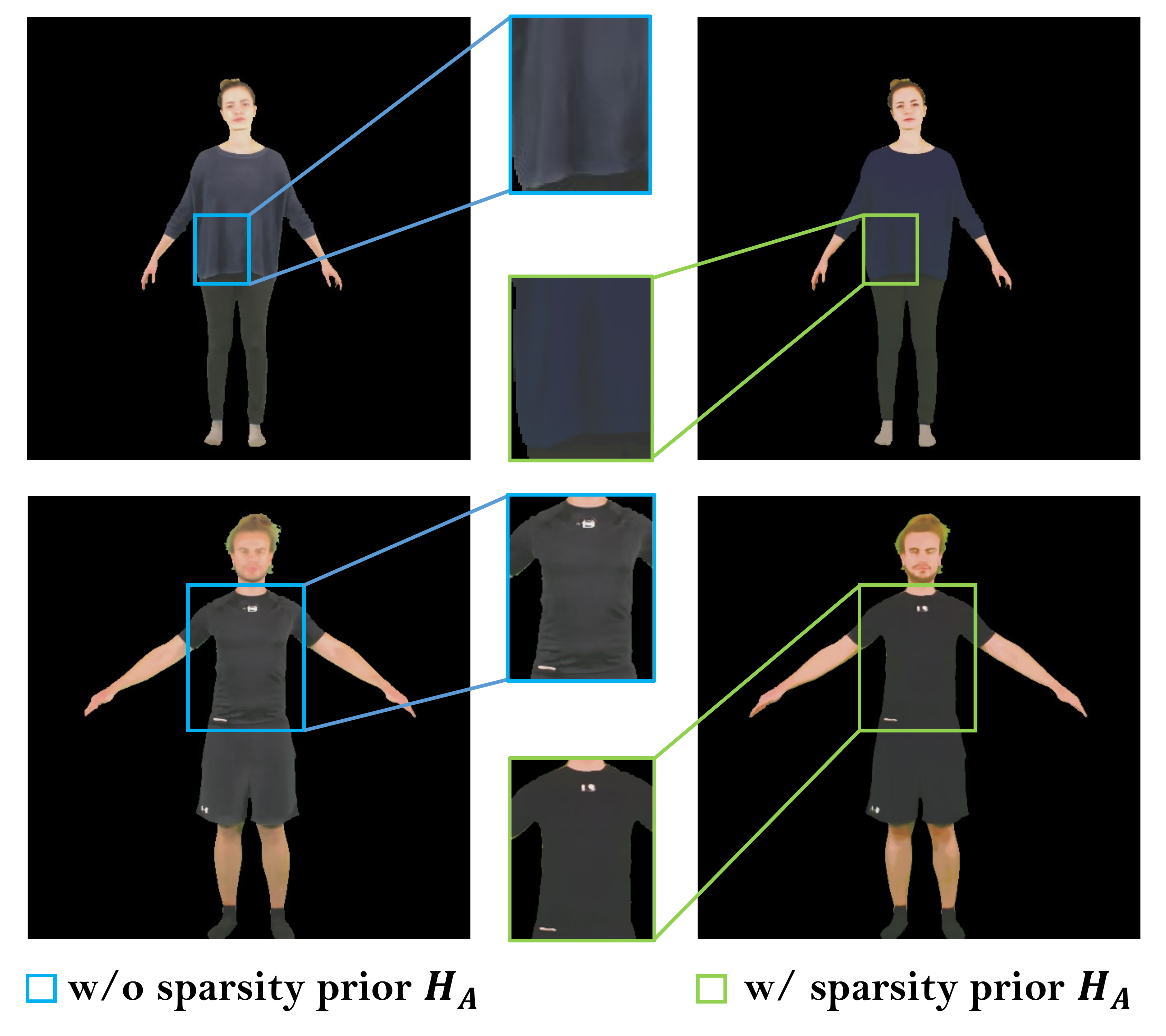}}
    \vskip -0.1in
    \caption{Here we visualize how our model benefits from incorporating minimum-entropy sparsity prior by minimizing $H_A$. Without this prior, the estimation of diffuse map would suffer from shadow residuals as shown on the left side. 
    }\label{fig:albedo-entropy}
    \end{minipage}
    \end{center}
\end{figure}

We demonstrate that our method is capable of relighting dynamic humans with complex motions from multi-views videos on the ZJU-Mocap dataset~\cite{peng_neural_2021}. Figure \ref{fig:vis-zju} shows our qualitative results on the "Twirl" and "Swing" scenes.

\begin{figure}[t]
    \begin{center}
    \begin{minipage}{0.50\columnwidth}
    \centerline{\includegraphics[width=1.0\columnwidth]{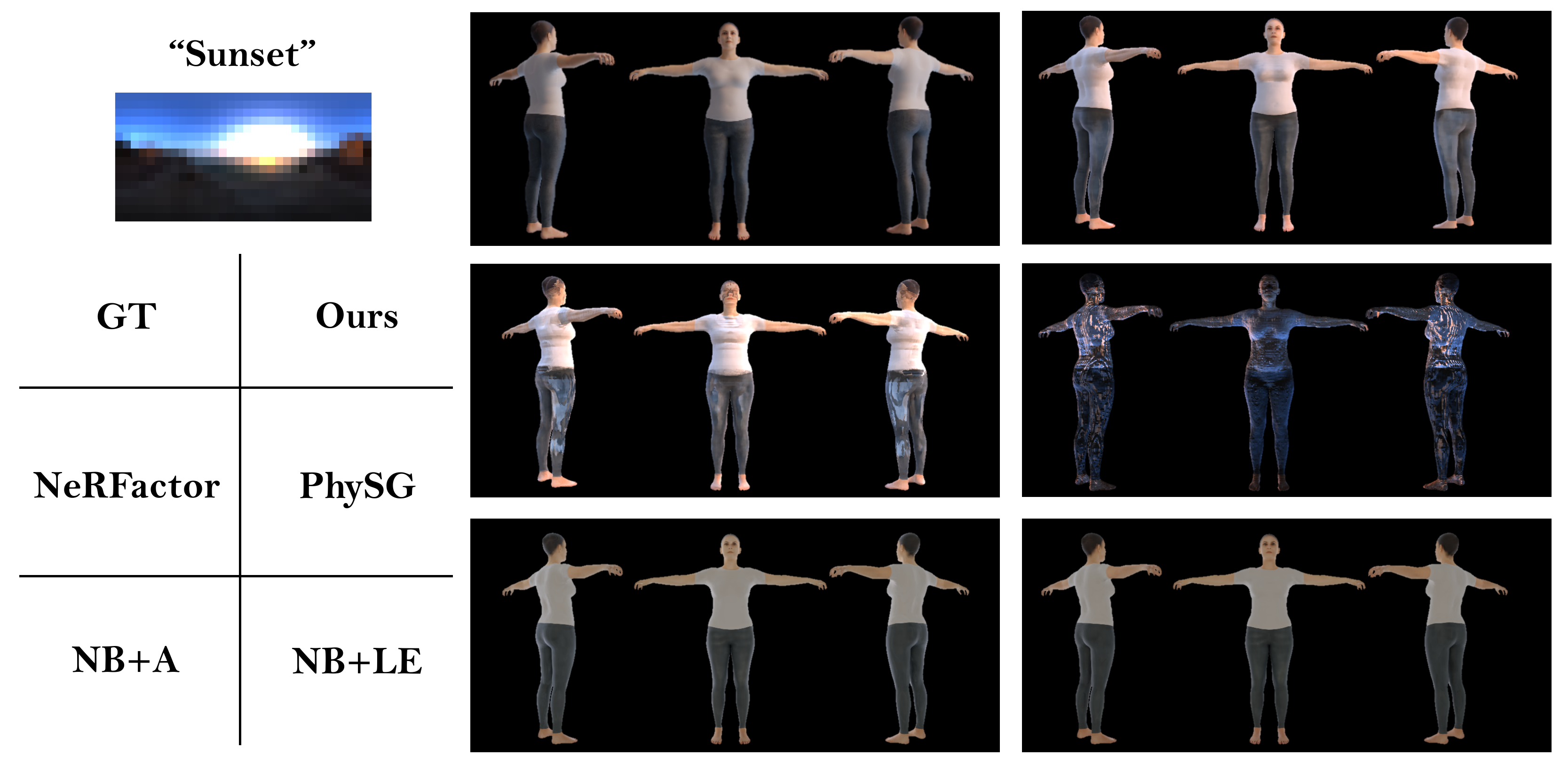}}
    \vskip 0.12in
    \caption{\textbf{Visualization of the "Sunset" scene in the BlenderHuman dataset}. We customize a synthetic dataset, BlenderHuman, to provide ground truths of relit videos for quantitative evaluations. \framework \; outperform other methods, producing promising results of relighting dynamic human under novel illuminations.
    }\label{fig:blender-vis-main}
    \end{minipage}
    \hspace{0.01in}
    \begin{minipage}{0.48\columnwidth}
    \centerline{\includegraphics[width=1.0\columnwidth]{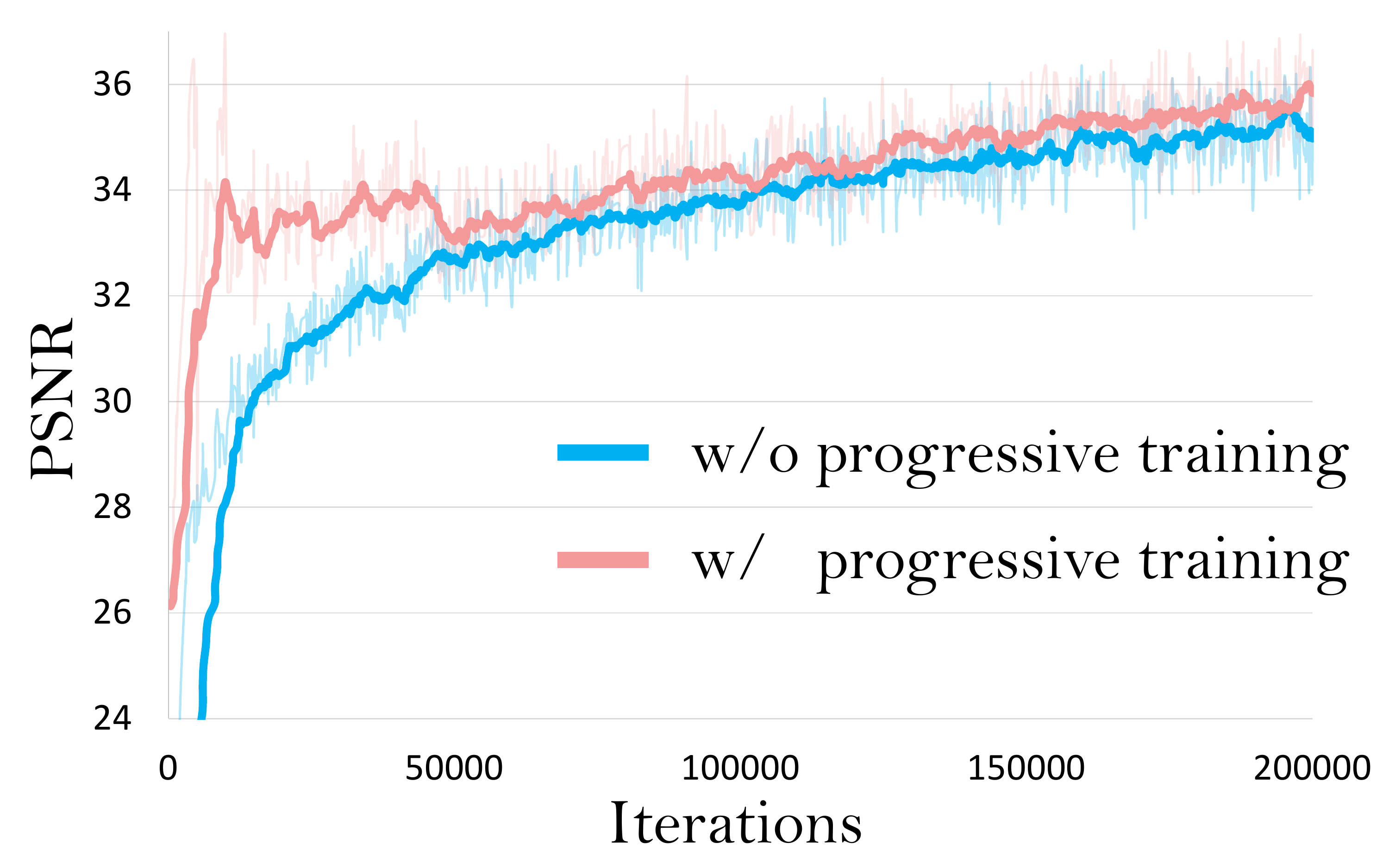}}
    \vskip -0.1in
    \caption{\textbf{The PSNR v.s. training iterations on People-Snapshot dataset.} Progressive training helps the model reconstruct the scene faster and better. When trained with a constant spatial resolution, the reconstruction error falls into sub-optimal at the end (PSNR drops from 36.65 to 34.56).
    }\label{fig:abl-pyramid}
    \end{minipage}
    \end{center}
\end{figure}

\begin{figure}[t]
    \centerline{\includegraphics[width=1.0\columnwidth]{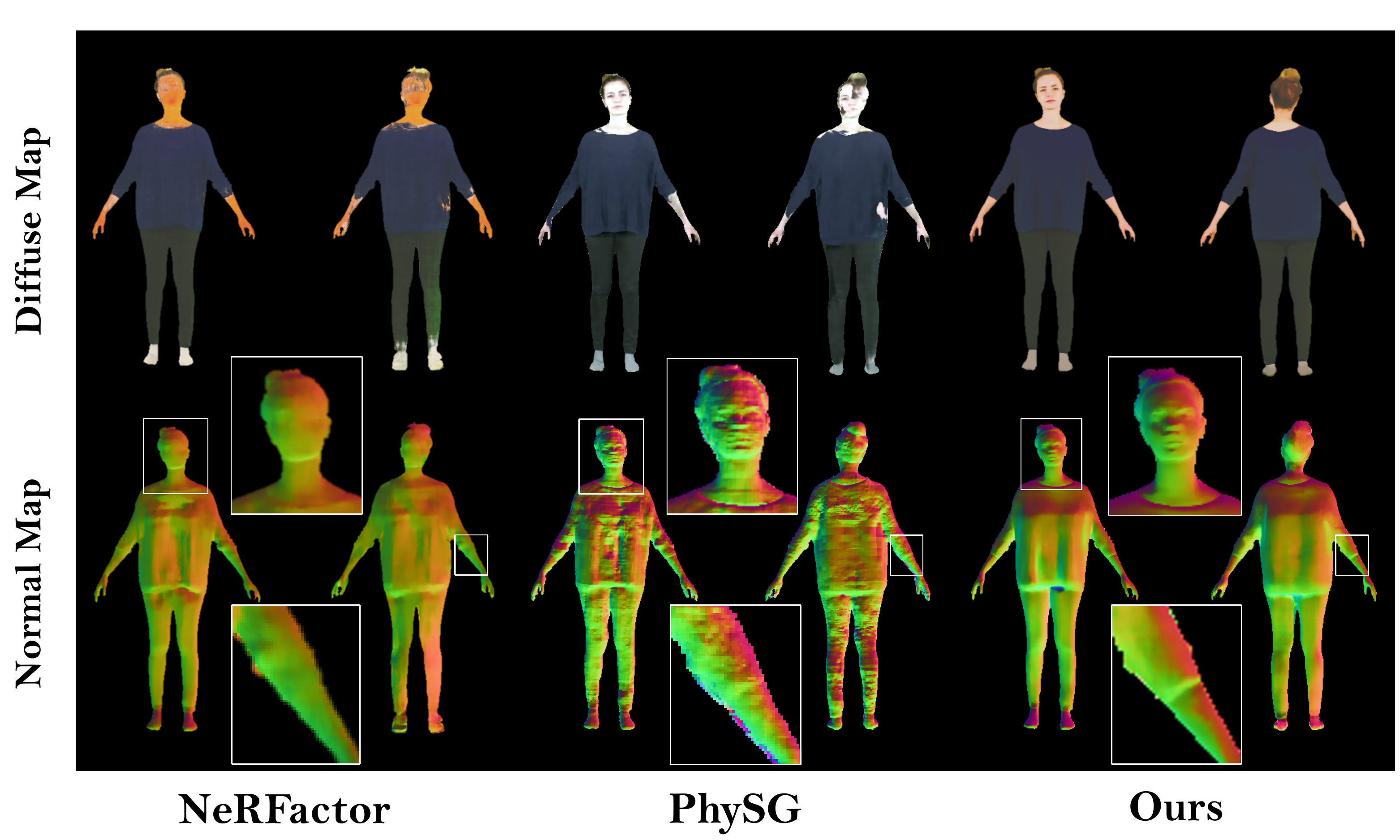}}
    \caption{\textbf{Comparisons of geometry and reflectance decomposition}. \framework \;is able to estimate fine-grained details of geometry and physically correct reflectance. We defer the visualization of the occlusion map in the supplementary.
    }\label{fig:disentanglement}
\end{figure}

\noindent\textbf{Decomposition of geometry and reflectance.} We demonstrate that our method is able to extract geometry and reflectance representations from the input videos and disentangle them into surface normals, diffuse maps and occlusion maps, which may facilitate downstream graphics tasks. The visualizations on People-Snapshot videos are presented in Figure \ref{fig:disentanglement}, and quantitative results on the BlenderHuman dataset are shown in Table \ref{tab:blenderhuman}. Note that, we directly take the albedo generated by NeRFactor~\cite{zhang_nerfactor_2021} and PhySG~\cite{zhang_physg_2021} as their diffuse maps for comparisons.
However, NeRFactor~\cite{zhang_nerfactor_2021} estimates the diffuse map of the dynamic human with incorrect base color and facial details, and fails to capture the accurate geometry of dynamic humans. Though PhySG~\cite{zhang_physg_2021} captures the correct base color of clothes, due to its incapability of handling dynamic scenes, the facial details of diffuse map remains artifacts when the viewpoint changes. With the latent representation of human body, \framework \;can integrate geometry information through space and time, successfully capturing the fine-grained details of the normal map and the correct color of diffuse map.

\subsection{Results on Synthetic Dataset}
\label{sec:exp-blender}

We quantitatively evaluate comparison methods on the BlenderHuman dataset. We show results in this section and defer the visualizations in the supplementary.

\begin{table*}[!t]
    \centering
    \begin{footnotesize}
    \caption{\textbf{Results on the BlenderHuman dataset}. The top two techniques for each metric are highlighted in \textcolor{rred}{red} and \textcolor{oorange}{orange} respectively. The reported numbers are the arithmetic averages of 16 different scenes. We relight the human actor with 8 HDR ambient light probes and 8 OLAT conditions. \framework \;achieves the best overall performance across all metrics. }
    \label{tab:blenderhuman}
    \setlength{\tabcolsep}{2mm}{
    \vspace{5pt}
    \renewcommand\arraystretch{0.975}
    \resizebox{1.0\columnwidth}{!}{
    \begin{tabular}{cccccccc} 
    \toprule
    \multirow{2}{*}{Method}& \multicolumn{3}{c}{\textbf{Relighting}}  & \multicolumn{1}{c}{\textbf{Normal Map}}  & \multicolumn{3}{c}{\textbf{Diffuse Map}}\\ 
    & \footnotesize{PSNR\;$\uparrow$} & \footnotesize{SSIM\;$\uparrow$} & \footnotesize{LPIPS\;$\downarrow$} & \footnotesize{Degree$^\circ\downarrow$} & \footnotesize{PSNR\;$\uparrow$} & \footnotesize{SSIM\;$\uparrow$} & \footnotesize{LPIPS\;$\downarrow$}\\  
    \midrule
    NB\cite{peng_neural_2021}+A\;\;&20.9348&0.8559&0.2368&-&-&-&-\\
    NB\cite{peng_neural_2021}+LE&22.7957&0.8721&0.2145&-&-&-&-\\
    NeRFactor\cite{zhang_nerfactor_2021}&22.8037&\cellcolor{oorange}0.8830&\cellcolor{oorange}0.2045&\cellcolor{oorange}43.7012&27.0585&0.9202&0.1929\\
    PhySG\cite{zhang_physg_2021}&\cellcolor{oorange}23.8810&0.8427&0.2959&50.5721&\cellcolor{oorange}28.0852&\cellcolor{rred}0.9350&\cellcolor{oorange}0.1810
\\
    Ours&\cellcolor{rred}26.1475&\cellcolor{rred}0.9118&\cellcolor{rred}0.1639&\cellcolor{rred}32.1803&\cellcolor{rred}28.9517&\cellcolor{oorange}0.9279&\cellcolor{rred}0.1502\\
    \bottomrule
    \end{tabular}}}
    \end{footnotesize}
\end{table*}

Table \ref{tab:blenderhuman} shows the results on the BlenderHuman dataset. Overall, our model achieves the best performance. The results indicate that \framework \;better handles the dynamics across video frames while feasibly modeling the light transport to relight dynamic humans.

\subsection{Ablation Studies}
\label{sec:exp-abl}

We conduct ablation studies on BlenderHuman dataset, as presented in Table \ref{tab:abl}.

\noindent\textbf{Impact of the human representation.} We train our model without $\psi_t(\boldsymbol{x})$ that is proposed in Section \ref{sec:latent}. In other words, the geometry and reflectance MLPs take only the surface coordinates of the human body as inputs. The result indicates that incorporating our latent feature  $\psi_t(\boldsymbol{x})$ is crucial for relighting dynamic human videos. The incapability of modeling scene dynamics leads to significant performance drops in terms of relighting quality and inverse rendering quality, which explains why NeRFactor~\cite{zhang_nerfactor_2021} fails so badly.

\noindent\textbf{Effectiveness of the smoothness regularization} is validated in Table \ref{tab:abl}. We train our model without $\tau_V, \tau_n$, which leads to the decreased rendering quality.

\noindent\textbf{Impact of the baked geometry.} We train \framework \;without the supervision of the baked geometry. The results indicate that the baked geometry improves the relighting performance. It reveals that poor renderings are caused by the inaccurate geometry as the error of normals drops by a large margin.

\noindent\textbf{Effectiveness of the global sparsity prior} on diffuse map is validated in Table \ref{tab:abl}. Without minimizing the entropy of the diffuse map, the relighting quality is perceptually decreased due to the degraded inverse rendering, which induces shadows in the estimation of diffuse map, as shown in Figure \ref{fig:albedo-entropy}.

\noindent\textbf{Impact of progressive training}. Without progressive spatial resolutions during training, the relighting quality decreases as shown in the last row of Table \ref{tab:abl}. We believe the progressive strategy helps the model quickly learn coarse geometry in the early training phase, which is even validated on real datasets (Figure \ref{fig:abl-pyramid}). We plot the reconstruction error (PSNR) versus iterations on one training scene in People-Snapshot dataset, discovering that progressive training helps model reconstruct the given scene faster and better.

\begin{table*}[!t]
    \centering
    \begin{footnotesize}
    \caption{\textbf{Ablation studies on the BlenderHuman dataset}. We take the average metrics on all 16 scenes. The top three techniques for each metric are highlighted in \textcolor{rred}{red} , \textcolor{oorange}{orange}, and \textcolor{yyellow}{yellow} respectively.}\label{tab:abl}
    \setlength{\tabcolsep}{2mm}{
    \vspace{5pt}
    \renewcommand\arraystretch{0.975}
    \resizebox{1.0\columnwidth}{!}{
    \begin{tabular}{cccccccc} 
    \toprule
    \multirow{2}{*}{Method}& \multicolumn{3}{c}{\textbf{Relighting}}  & \multicolumn{1}{c}{\textbf{Normal Map}}  & \multicolumn{3}{c}{\textbf{Diffuse Map}}\\ 
    & \footnotesize{PSNR\;$\uparrow$} & \footnotesize{SSIM\;$\uparrow$} & \footnotesize{LPIPS\;$\downarrow$} & \footnotesize{Degree$^\circ\downarrow$} & \footnotesize{PSNR\;$\uparrow$} & \footnotesize{SSIM\;$\uparrow$} & \footnotesize{LPIPS\;$\downarrow$}\\  
    \midrule
    Full model&\cellcolor{oorange}26.1475&\cellcolor{rred}0.9118&\cellcolor{rred}0.1639&\cellcolor{yyellow}32.1803&\cellcolor{rred}28.9517&\cellcolor{rred}0.9279&\cellcolor{oorange}0.1502\\
    w/o $\psi_t(\boldsymbol{x})$&21.1163&0.8407&0.2372&36.5699&25.6806&0.9008&0.1896\\
    w/o $\tau_V, \tau_n$&25.3504&0.8800&0.2061&32.9243&\cellcolor{oorange}28.3329&\cellcolor{oorange}0.9224&\cellcolor{yyellow}0.1660\\
    w/o $\Tilde{\boldsymbol{V}}, \Tilde{\boldsymbol{n}}$&22.4221&0.8559&0.2285&57.0452&\cellcolor{yyellow}27.6652&\cellcolor{yyellow}0.9165&\cellcolor{rred}0.1425\\
    w/o $H_A$&\cellcolor{rred}27.7545&\cellcolor{oorange}0.9042&\cellcolor{oorange}0.1717&\cellcolor{oorange}30.6685&24.0195&0.8950&0.1767\\
    w/o progressive&\cellcolor{yyellow}25.5562&\cellcolor{yyellow}0.9031&\cellcolor{yyellow}0.1742&\cellcolor{rred}30.1662&24.2455&0.8958&0.1760\\
    \bottomrule
    \end{tabular}}}
    \end{footnotesize}
\end{table*}

\section{Discussion and Conclusion}

\noindent\textbf{Limitations.} We have demonstrated the capability of \framework \;on relighting dynamic humans with free viewpoints. Nevertheless, there are a few limitations. 
First, for tractable training and rendering, we consider only the one-bounce direct environment light, thus our method cannot relight furry appearances. 
Second, as we leverage a fully physically based renderer, the rendering quality is tied with the accuracy of geometry.
Dense scenes with multiple people, which may negatively impact the estimation of geometry, will lead to poor performance. 
Finally, if the texture patterns are complicated or the lighting is harsh during training, the decomposition of reflectance and geometry is hard to solve due to the ambiguity of color scale, causing poor relighting quality. It can be alleviated by incorporating more information other than self-supervision from videos into the network (e.g., other supervision signals or data-driven priors).

In this paper, we present a principled rendering scheme called \framework, a method that enables relighting with free viewpoints from only posed human videos under unknown illuminations. Our method exploits the physically based rendering pipeline and decomposes the appearance of humans into geometry and reflectance. All components are parameterized by MLPs based on the neural field conditioned on the deformable human model. Extensive experiments on synthetic and real datasets demonstrate that \framework \;is capable of high-quality relighting of dynamic human performers with free viewpoints. \\

\noindent \textbf{Acknowledgements} \quad This work is supported by the National Research Foundation, Singapore under its AI Singapore Programme (AISG Award No: AISG2-PhD-2021-08-019), NTU NAP, MOE AcRF Tier 2 (T2EP20221-0033), and under the RIE2020 Industry Alignment Fund - Industry Collaboration Projects (IAF-ICP) Funding Initiative, as well as cash and in-kind contribution from the industry partner(s).
%
%
\newpage
\bibliographystyle{splncs04}
\bibliography{egbib}
\end{document}